\documentclass[letterpaper, 10 pt, conference]{ieeeconf}

\usepackage{etoolbox}
\makeatletter
\patchcmd{\@makecaption}
  {\scshape}
  {}
  {}
  {}
\makeatletter
\patchcmd{\@makecaption}
  {\\}
  {.\ }
  {}
  {}
\makeatother
\def\tablename{Table}

\IEEEoverridecommandlockouts                              
\usepackage{times}

\usepackage{multicol}
\usepackage[bookmarks=true]{hyperref}
\usepackage{graphicx}
\usepackage{tabularx}
\usepackage{amssymb}
\usepackage{amsmath}
\usepackage{fancyhdr}
\usepackage{tikz}
\usepackage{bm}
\usepackage{listings}
\usepackage{subfigure}
\usepackage{xcolor}
\usepackage{scalerel}
\usepackage[colorinlistoftodos]{todonotes}
\usepackage{algorithm}
\usepackage[noend]{algpseudocode}
\usepackage{algorithmicx}
\usepackage{float}
\usepackage{tikz}
\usepackage{graphics} 
\usetikzlibrary{positioning}
\usetikzlibrary{shapes,arrows}
\usepackage{dcolumn}
\usepackage{multirow}
\usepackage{makecell}
\usepackage[font=small,skip=2pt]{caption}

\newcommand{\vectg}[1]{\boldsymbol{#1}}

\newcommand{\bx}{\mathbf{x}}

\newcommand{\bu}{\mathbf{u}}

\newcommand{\bp}{\mathbf{p}}

\newcommand{\bX}{\boldsymbol{\tau}}
\newcommand{\bU}{\boldsymbol{\xi}}
\newcommand{\bK}{\mathbf{K}}

\newcommand{\bnu}{\boldsymbol{\nu}}

\newcommand{\bmu}{\boldsymbol{\mu}}

\newcommand{\bSigma}{\boldsymbol{\Sigma}}
\newcommand{\bff}{\mathbf{f}}

\DeclareMathOperator*{\argmin}{arg\,min}

\def\NAT@parse{\typeout{IEEEtran error: Attempt to use fake Natbib command 
which is provided to fool Hyperref.}}

\allowdisplaybreaks
\renewcommand{\baselinestretch}{0.98}
\begin{document}

\onecolumn
\newpage
\thispagestyle{empty}  
\begin{center}
    \vspace*{\fill}  
    \textcopyright 2025 IEEE. Personal use of this material is permitted. Permission from IEEE must be obtained for all\\
    other uses, in any current or future media, including reprinting/republishing this material for advertising\\
    or promotional purposes, creating new collective works, for resale or redistribution to servers or lists,\\
    or reuse of any copyrighted component of this work in other works.
    \vspace*{\fill}  
\end{center}

\newpage 
\twocolumn

\title{\LARGE \bf Multi-Agent Feedback Motion Planning using Probably Approximately Correct Nonlinear Model Predictive Control}
\author{Mark Gonzales$^{1}$, Adam Polevoy$^{1,2}$, Marin Kobilarov$^{1}$, Joseph Moore$^{1, 2}$
\thanks{$^{1}$Johns Hopkins University \newline \hspace*{1.6em}
{\tt\small \{MGonza60,\newline \hspace*{1.6em} Marin\}jhu.edu} 
\newline \hspace*{0.8em} $^{2}$Johns Hopkins University Applied Physics Lab \newline \hspace*{1.6em} {\tt\small \{Adam.Polevoy,\newline \hspace*{1.6em} Joseph.Moore\}@jhuapl.edu}}}
\maketitle

\begin{abstract}

For many tasks, multi-robot teams often provide greater efficiency, robustness, and resiliency. However, multi-robot collaboration in real-world scenarios poses a number of major challenges, especially when dynamic robots must balance competing objectives like formation control and obstacle avoidance in the presence of stochastic dynamics and sensor uncertainty. In this paper, we propose a distributed, multi-agent receding-horizon feedback motion planning approach using Probably Approximately Correct Nonlinear Model Predictive Control (PAC-NMPC) that is able to reason about both model and measurement uncertainty to achieve robust multi-agent formation control while navigating cluttered obstacle fields and avoiding inter-robot collisions. Our approach relies not only on the underlying PAC-NMPC algorithm but also on a terminal cost-function derived from gyroscopic obstacle avoidance. Through numerical simulation, we show that our distributed approach performs on par with a centralized formulation, that it offers improved performance in the case of significant measurement noise, and that it can scale to more complex dynamical systems.

\end{abstract}
\IEEEpeerreviewmaketitle

\section{Introduction}
Unified control and path planning for multi-robot systems is a challenging problem to address, especially in the presence of complex stochastic dynamics and measurement uncertainty. Even in discrete state and action spaces, the complexity of finding an optimal solution to the multi-agent path planning problem is NP-hard \cite{stern2019multi}, and the computation time increases exponentially with the number of robots. The computational complexity is only further exacerbated in more realistic scenarios characterized by continuous state and action spaces where the multi-robot team must achieve objectives like formation control, obstacle avoidance, and reason about stochastic underactuated nonlinear dynamics.

Centralized controllers, as referenced in \cite{Sandip2012}, \cite{Wu2021}, \cite{Beyoglu2022},  and \cite{Liu2017} are commonly employed to address these issues. However, these centralized approaches can be slow and not scale as the team size increases. Another limitation is that centralized approaches may not be able to update quickly enough to accommodate changing environments or imperfect environmental data, which could compromise the effectiveness of the path plan.

This paper builds on the sampling-based Stochastic Nonlinear Model Predictive Control (SNMPC) algorithm, Probably Approximate Correct NMPC (PAC-NMPC), as cited in \cite{Polevoy2023}, to enable distributed multiple robot collaboration and feedback motion planning in the presence of static and dynamic obstacles. In particular, we introduce a set of costs and constraints that capture the expected behavior of teammates over a finite horizon and enable probabilistically-safe formation control in cluttered environments.
Our contributions are:
\begin{itemize} 
    \item A receding-horizon feedback motion planning framework for distributed, probabilistically-safe multi-robot formation control in obstacle fields under dynamics and measurement uncertainty.
    \item A terminal cost inspired by gyroscopic obstacle avoidance to improve finite-horizon planning in the presence of static and dynamic obstacles.
    \item The demonstration of the algorithm’s effectiveness through numerical simulation in complex environments and its ability to scale to higher dimensions.
\end{itemize}

\begin{figure}[t]
    \centering
    \includegraphics[width=\columnwidth]{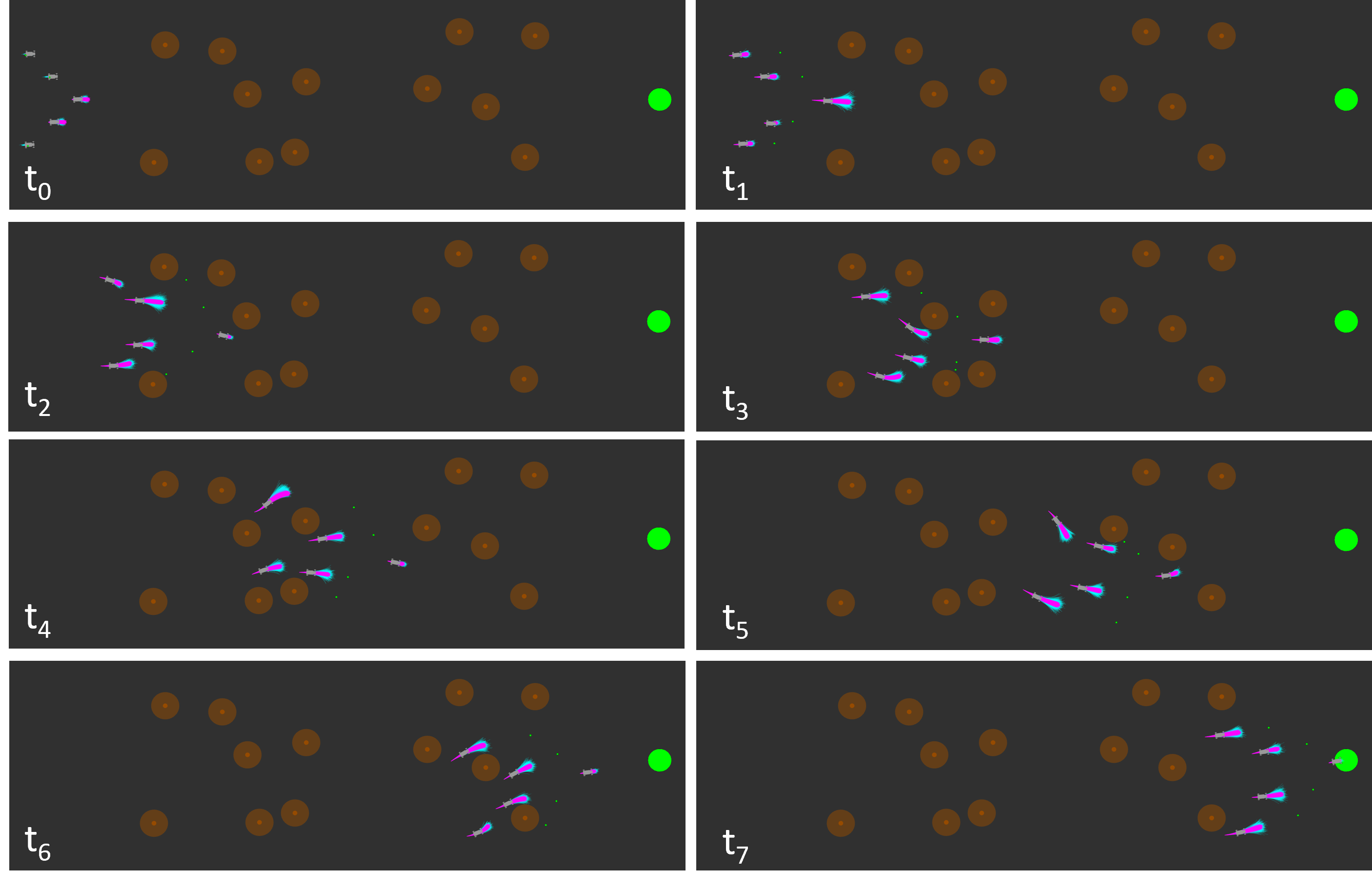}
    \caption{A five-agent team in a wedge formation in an obstacle field.}
    \label{fig:Wedge}
    \vspace{-15pt}
\end{figure}


\section{Related Work}




Researchers have explored a wide variety of strategies for achieving unified formation control and obstacle avoidance \cite{Hai2021} \cite{Chen_2023}. Behavior-based controllers are one class of methods that have tried to address this problem \cite{Balch_1998},\cite{Lee_2018}, \cite{Emile_2020}. These controllers combine the outputs from a set of sub-controllers that promote separate objectives, such as formation error, obstacle avoidance, and goal error. Typically, these approaches suffer from poor robustness and generalizability.

Consensus-based approaches achieve formation control by relying on local information to facilitate the convergence of all agents to the same information state\cite{ren2006consensus,Xiao_2009}. Some approaches have explored the joint formation control and obstacle avoidance problem \cite{saber2003flocking} and others have extended consensus-based formation control to simple non-holonomic systems (e.g., \cite{Wang_2014}, \cite{Romero_2023}). 

Artificial Potential Fields (APF) are a set of approaches stemming from the path planning approach by Khatib \cite{Khatib1990}. APFs have been used to achieve formations while avoiding obstacles \cite{Colledanchise_2013} \cite{Liu_2017} \cite{Sun_2020}. These allow for fixed formation and re-updating the controller after every time step and can be used for both leader-follower and other formation types, but are often limited to simple dynamics models. 



Virtual structures achieve formation control by reasoning about the geometric performance of a rigid structure \cite{Tan_1996} \cite{Lewis_1997}. Virtual structures have been used to adapt Probabilistic Roadmaps (PRM) \cite{Wu_2021}. Typically, virtual structures offer limited flexibility and obstacle avoidance.




Machine Learning and Reinforcement Learning have been used to handle multi-agent formation. For swarm robotics, learning has been used on the dynamics of nonlinear systems to give bounds on formation error \cite{Beckers_2021}. Other approaches use policy search to solve nonlinear optimization problems within formation control \cite{Lima_2021}  \cite{Ryou_2022}. Reinforcement learning for leader-follower and swarm robotics has been used, training Double Deep Q-Networks to obtain proper formation behavior \cite{Obradović_2023} or to achieve multi-robot planning in the presence of dynamic obstacles \cite{Xue_2021}. Oftentimes, learning-based methods demonstrate degraded performance in out-of-distribution environments and during sim-to-real transfer.


In recent years, online trajectory optimization and Nonlinear Model Predictive Control (NMPC) have emerged a powerful approaches for achieving multi-robot planning and control. Some approaches have applied NMPC to achieve formation control \cite{Roldao_2014,Xiao_2016,Xiao_2017,Xiao_2019,Xu_2023}. Other approaches have explored inter-agent collision avoidance via distributed NMPC \cite{Luis_2020,Trevisan2024} or by combining NMPC with conflict-based search \cite{tajbakhsh2024conflict}. A few approaches have explored NMPC to achieve formation control in the presence of obstacles. \cite{Nfaileh_2022} employs a virtual structure framework and \cite{Satir_2023} is restricted to a linear dynamics model.  

To our knowledge, our approach is the first method to employ SNMPC to achieve distributed, probabilistically-safe multi-agent collision avoidance and formation control in cluttered environments for a large class of stochastic nonlinear dynamical systems and measurement uncertainty.

\section{Probably Approximately Correct Nonlinear Model Predictive Control (PAC-NMPC)}
Our proposed multi-agent formation controller utilizes Probably Approximately Correct Nonlinear Model Predictive Control (PAC-NMPC) \cite{Polevoy2023}. PAC-NMPC uses Iterative Stochastic Policy Optimization (ISPO) \cite{kobilarov2015sample} to formulate the search for a local time-varying feedback control policy of the form $\bu_t =\bK_t(\bX^d(\bU,\bx_0))\left(\bx_t^d(\bU,\bx_0) - \bx_t\right) + \bu_t^d(\bU)$, as a stochastic optimization problem. This is achieved by sampling the policy parameters, $\bU$, from a surrogate distribution given by the probability density function $p(\bU|\bnu)$ and defined by hyper-parameters $\bnu$. Here $\bX^d\triangleq\{\bx_0^d,\bu_0^d, \bx_1^d,\bu_1^d , ... , \bu_{N_T-1}^d, \bx_{N_T}^d\}$ is the nominal trajectory computed using a nominal deterministic discrete-time dynamics model
$\bx_{t+1}^d=\bx_t^d+\bff(\bx_t^d,\bu_t^d)\Delta t$
over $N_T$ time steps. $\bx_t^d\in \mathbb{R}^{N_x}$ and $\bu_t^d\in \mathbb{R}^{N_u}$ are the nominal states and control inputs at time index $t$ respectively. $\Delta t$ is the discrete time step and $\bK_t(\bX^d(\bU,\bx_0))\in\mathbb{R}^{N_u\times N_x}$ is a sequence of time-varying feedback gains computed using the finite horizon, discrete, time-varying linear quadratic regulator (TVLQR) \cite{underactuated}. The policy is parameterized only by the nominal input sequence, so that $\bU ~= [ {\bu^d_0}^T \ {\bu^d_1}^T \ ... \ {\bu^d_{N_T-1}}^T]^T$. The surrogate distribution,  $p(\bU|\bnu)$, is parameterized as a multivariate Gaussian, $\mathcal{N}\left(\bU | \bmu, \bSigma \right)$, with a mean, $\bmu$, and a covariance matrix, $\bSigma$. Since the covariance matrix is diagonal, the hyperparameters can be written as $\bnu \triangleq [\bmu^T \ diag(\bSigma)^T]^T$. For a discrete-time trajectory sequence $\bX=\{\bx_0,\bu_0, \bx_1,\bu_1 ..., \bu_{N_T-1}, \bx_{N_T} \}$, one can define a non-negative trajectory cost function, $J(\bX) \ge 0$, and a trajectory constraint violation function $C(\bX) \in \{0, 1\}$.

During each planning iteration, PAC-NMPC optimizes a weighted objective function $\bnu^* = \argmin_{\bnu} \min_{\alpha > 0} (\mathcal{J}^+_\alpha(\bnu) + \gamma \mathcal{C}^+_\alpha(\bnu))$
where $\mathcal{J}^+_\alpha(\bnu)$ is the PAC bound on $J(\bX)$, $\mathcal{C}^+_\alpha(\bnu)$ is the PAC bound on $C(\bX)$, and $\gamma$ is positive weighting coefficient.  These PAC bounds, which are derived in \cite{kobilarov2015sample}, take the form
\begin{align}
&\mathcal{J}^+_\alpha(\vectg{\nu})\! \triangleq \! \widehat{\mathcal J}_\alpha(\vectg{\nu}) + \alpha d(\vectg{\nu}) + \Phi_{\alpha}(\delta), 
\label{eq:pacbound}
\end{align}
where $\widehat{\mathcal J}_\alpha(\vectg{\nu})$ is a robust estimator of the expected cost, $d(\vectg{\nu})$ is a distance between distributions, $\Phi_{\alpha}(\delta)$ is a concentration-of-measure term, and $1 - \delta$ is the bound confidence.

This optimization ensures that the chosen control policy satisfies performance and safety requirements with a specified confidence level. In essence, PAC-NMPC offers a robust and statistically guaranteed approach for controlling systems with uncertainties.

\section{PAC-NMPC for Multi-Agent Control}
This section presents a centralized and distributed approach for PAC-NMPC to control a team of $M$ agents in formation, navigating an obstacle-filled environment to a goal state $\bx^G$. 

\subsection{Problem Formulation}
For the centralized approach for formation control, we define the team state as $\bx = [\bx^{1T}, \bx^{2T}, \bx^{3T}, ..., \bx^{MT}]^T$ where $\bx^i$ is the state of the $i^{th}$ agent on the team.
To guide the team towards $\bx^G$, we define a cost function $J(\bX)$ for the team's trajectory as
\begin{align}
J(\bX) = \sum_{t=0}^{N_T-1} q(\bx_t, \bu_t) + q_f(\bx_{N_T})
\end{align}
where $q(\bx_t, \bu_t)$ is the cost at timestep $t$ and $q_f(\bx_{N_T})$ is the cost at the final time. Constraints to ensure collision avoidance and bound the state are formulated as
\begin{align}
g_b(\bx_t) &= (\bx_t - \bx_l <0) \lor (\bx_u - \bx_t)  <0\\
g_{o}(\bx_t) &= \{dist(\bx_t, \bp^{o^m}) - r \} \ \forall \ m\nonumber\\
c(\bx_t) &= g_b(\bx_t) < 0 \lor g_{o}(\bx_t) < 0 \lor g_{A}(\bx_t) < 0 \nonumber\\
C(\bX) &= c(\bx_0) \lor c(\bx_1) \cdots \lor c(\bx_{N_T})\nonumber.
\end{align}
Here $\bx_{l,u}$ is the lower and upper state bound, $r$ is the obstacle radius plus the maximum robot radius, $\bp^{o^m}$ is the $m^{th}$ obstacle position in the world frame. 
$g_A$ is a constraint to prevent collisions between dynamic agents and is formulated as
\begin{align}
g_a(\bx_t^i) &= \{dist(\bx_t^i, \bx_t^n) - L \} \ \forall \ n: i \neq n\\
g_A(\bx_t) &= g_a(\bx_t^1) \lor g_a(\bx_t^2) \cdots \lor g_a(\bx_t^{M})\nonumber
\end{align}
where $L$ is the collision radius and $\bx_t^n$ is $n^{th}$ agent on the team.





\subsection{Cost Function}
We formulate our cost function for the $i^{th}$ agent as
\begin{align}
J^i(\bX)=\sum_{t=0}^{N_T-1}\big(\omega_1(||k\tilde{\bp}^i_t|| - ||\mathbf{v}^i_t||)^2 + \omega_3||\bu_t^i||^2\big) + \tilde{\bf v}^{iT}{\bf Q}_f \tilde{\bf v}^i
\end{align}
where $\tilde{\bp}_t^i = \bp_t^i-\bp^{G,i}$, $\bp_t^i$ is the position of agent $i$ at time $t$, and $\bp^{G,i}$ is the final desired position of agent $i$. $\tilde{\bf v}^i = {\bf v}_d^i-{\bf v}_{N_T}^i$, where ${\bf v}_d^i$ is the desired velocity vector for the terminal state of agent $i$, and ${\bf v}_{N_T}^i$ is the final velocity of agent $i$. The first term in the running cost encourages a desired velocity, while the second term applies a cost on action. The terminal cost includes ${\bf v}_d^i$, which is a desired final velocity vector inspired by the gyroscopic obstacle avoidance control in \cite{Garimella2016}. It is given as
\begin{align}
\mathbf{v}_d^i = -(\sum_j \mathbf{G}(\theta^{ij}, {\bf d}^{ij}, \tilde{\bp}_{N_T}^i,{\bf v}^j) + {\bf I})k\tilde{\bp}_{N_T}^i 
\end{align}
where $||\mathbf{v}_d^i|| \leq \mathbf{v}_{max}$,
\begin{align}
&\mathbf{G}(\theta^{ij}, {\bf d}^{ij}, \tilde{\bp}_{N_T}^i,{\bf v}^j) = k_1(\theta^{ij})k_2(\mathbf{d}^{ij})e(\mathbf{d}^{ij},\tilde{\bp}_{N_T}^i,{\bf v}^j)\hat{\mathbf{e}},
\end{align}
and
\begin{align}
     & k_1(\theta^{ij}) = \exp(k_{att}(\theta^{ij} - 1)), \\
        & k_2(\mathbf{d}^{ij}) = k_{obs}\frac{S(r_d + r - ||\mathbf{d}^{ij}|| - \epsilon)}{||\mathbf{d}^{ij}|| - r},S(s) = \frac{1}{1 + e^{-s}}\nonumber,\\
     &\theta^{ij} = \frac{-\mathbf{d}^Tk(\tilde{\bp}_{N_T})}{||\mathbf{d}_i||||-k(\tilde{\bp}_{N_T})||}, \mathbf{d}^{ij} = \bp_{N_T}^i-\bp^{c^j},\hat{\mathbf{e}} = \begin{pmatrix}
        0 & -1\\
        1 & 0
    \end{pmatrix}\nonumber
\end{align}
Where $k$ is a proportional feedback gain, $S(s)$ is a smooth sign function, $k_{att}$ is the attractive weighting, $k_{obs}$ is the avoidance weighting, $\mathbf{d}^{ij}$ is the distance to static or dynamic obstacle $j$, $r_d$ is the detection radius, $\epsilon$ is a tuning value for negligible obstacle gain at the detection radius and ${\bf v}_{max}$ is the maximum desired velocity. $\sum_i J^i(\bX)$ is the cost for the full multi-agent system. $\bp^{c^j}$ is the set of static and dynamic obstacle positions, which includes the set of static obstacles $\bp^{o^m}$ and the set of dynamic obstacles represented by the final positions of the other agents $\bp_{N_T}^{n},\forall n\neq i$. ${\bf v}^j$ is the velocity of $j^{th}$ obstacle. For dynamic obstacles (i.e., other agents), this is the terminal velocity ${\bf v}_{N_T}^n$, and for static obstacles, it is zero.

To handle collisions with both static and dynamic obstacles (i.e., other agents), we choose $e(\mathbf{d}^{ij},\tilde{\bp}_{N_T}^i,{\bf v}^j)$ based on \cite{chang2003} to maintain a consistent policy among team members for selecting a ``give-way'' direction. $e$ is given as follows:


\begin{itemize}
    \item[C1.] If $\mathbf{d}^{ij} \cdot \mathbf{v}^i \geq 0 \land \mathbf{d}^{ij} \cdot \mathbf{v}^j \geq 0$:
    \[
    e(\mathbf{d}^{ij},\mathbf{v}^i,\mathbf{v}^j) =
    \begin{cases} 
    1 & \text{if } \varphi(\mathbf{d}^{ij}, \mathbf{v}^i) - \varphi(\mathbf{d}^{ij}, \mathbf{v}^j) \geq 0 \\
    -1 & \text{otherwise}
    \end{cases}
    \]
    
    \item[C2.] If $\mathbf{d}^{ij} \cdot \mathbf{v}^i \geq 0 \land \mathbf{d}^{ij} \cdot \mathbf{v}^j < 0$: 
    \[
    e(\mathbf{d}^i,\mathbf{v}^i,\mathbf{v}^j) =
    \begin{cases} 
    1 & \text{if } \varphi(\mathbf{d}^{ij}, \mathbf{v}^i) - \varphi(\mathbf{v}^j, \mathbf{d}^{ij}) \geq 0 \\
    -1 & \text{otherwise}
    \end{cases}
    \]
    
    \item[C3.] If $\mathbf{d}^{ij} \cdot \mathbf{v}^i < 0 \land \mathbf{d}^{ij} \cdot \mathbf{v}^j < 0$: 
    \[
    e(\mathbf{d}^i,\mathbf{v}^i,\mathbf{v}^j) =
    \begin{cases} 
    1 & \text{if } \varphi(\mathbf{d}^{ij}, \mathbf{v}^i) - \varphi(\mathbf{d}^{ij}, \mathbf{v}^j) > 0 \\
    -1 & \text{otherwise}
    \end{cases}
    \]

    \item[C4.] Else:
    \[
    e(\mathbf{d}^{ij},\mathbf{v}^i,\mathbf{v}^j) = 0
    \]
    
\end{itemize}
where $\varphi(\cdot, \cdot)$ is the angle between two vectors, and ${\bf v}^i$ is defined as $-k(\tilde{\bp}_{N_T}^i)$

Gyroscopic obstacle avoidance was used over AFP or no obstacle avoidance because it helped avoid local minima more effectively and allowed for trajectory deconflicting.

\subsection{Distributed Multi-Agent PAC-NMPC}
Both the cost function and the constraints are dependent on the full multi-robot system trajectory, $\bX$. To enable a distributed control paradigm, the agents share their initial states, $\bx_0^i$, and control policy distribution parameters, $\bnu^i$, at the beginning of each planning interval. Each agent $i$ then optimizes its own control policy distribution, assuming the policies of the other agents remain fixed during the planning interval. To reconstruct the trajectory $\bX$ and evaluate the joint costs and constraints, each agent $i$ samples trajectory predictions $(\bX^j, \bU^j)$ for the other agents using the received policies and states.

For formation control, the formation points can be specified in the cost function as the desired terminal positions, $\bp^{G,i}$. In this paper, we explore several methods for computing these formation points.

\section{State Measurement Error}
In real-world applications, relying solely on shared state information leads to discrepancies due to sensor noise, communication delays, and environmental uncertainties. These factors introduce discrepancies between an agent's actual state and the state information received by other agents. Not accounting for the uncertainty in the state measurement could lead to collisions or worse performance.

We model the state measurement error as a zero-mean Gaussian noise with a covariance $\mathbf{\Sigma}_p$. The received state of agent $j$ is represented as $\hat{\bx}^j \sim \mathcal{N}(\bx^j,\mathbf{\Sigma}_p)$ where $\bx^j$ is the true state and $\hat{\bx}^j$ is the received state. To reason about state measurement uncertainty, the algorithm samples the state from the normal distributions for each agent during trajectory rollouts. This approach enables an agent to consider other agents' state uncertainties and plan its trajectory to enable robust, collision-free navigation under measurement uncertainty.




\section{Simulation Experiments}
We conducted Monte Carlo simulations to test the effectiveness of multi-agent PAC-NMPC for formation control and collision avoidance. 

We simulate a stochastic bicycle model with acceleration and steering rate inputs for each agent. We denote ${\bx_t}=[ p_x, p_y, \theta, v, \delta_s]^T$ as the state vector, ${\bu_t}=[\dot{v}, \dot{\delta_s}]^T$ as the control vector, $l = 0.33$ as the wheel base, and $\boldsymbol{\Gamma} = diag(\left[0.001, 0.001, 0.1, 0.2, 0.001\right])$ as the covariance. The stochastic bicycle model is defined as
\begin{align}
  \bx_{t+1} &\sim p(\cdot | \bx_t, \bu_t) \triangleq \bx_t + \left(f(\bx_t, \bu_t) + \boldsymbol{\omega} \right) \Delta t \\
  f(\bx_t, \bu_t) &= [v\cos(\theta), v\sin(\theta), \frac{v\tan(\delta_s)}{l}, \dot{v}, \dot{\delta_s}]^T \nonumber \\
  \vectg{\omega} &\sim \mathcal{N}(\cdot | \mathbf{0}, \boldsymbol{\Gamma})\nonumber
\end{align}

The acceleration is limited to $-1\ m/s^2 \leq \dot{v} \leq 1\ m/s^2 $, the steering rate is limited $-1\ rad/s \leq \dot{\delta_s} \leq 1\ rad/s$, the velocity is limited to $-0.5\ m/s  \leq v \leq 2\ m/s $, the steering angle is limited to $-0.4\ rad \leq \delta_s \leq 0.4\ rad$.


We generated fifty random obstacle fields of 10 non-overlapping circular obstacles with an inflated radius of 0.6 meters to test the formation control in obstacle-filled environments. The obstacles were between $ {\bf p}_{min} = [3,- 4] $ and $ {\bf p}_{max} = [10, 4] $. The fields were generated with free space before and after the obstacle field to allow the formation to converge. The desired formation was a 3-agent wedge. The formation goals were defined and calculated as
\begin{align}
    \bp^{G, 2,3} &= \bp^1 - \begin{pmatrix}
        \cos(\theta^1) & \pm sin(\theta^1)\\
        \sin(\theta^1) & \mp cos(\theta^1)
    \end{pmatrix}\begin{pmatrix}
        l\\h
    \end{pmatrix}
\end{align}

where $l$ is the distance in the x-direction behind the leader, and $h$ is the distance in the y-direction behind the leader. For these experiments, $l = h = 1 \text{ meter}$. We chose these distances to make a formation too small to wrap around obstacles but not large enough where the agents were too far away to interact.

We chose a goal at $\bp^{G,1}$ = [15, 0]. We sampled the leader agent state from a uniform distribution between $x_{min}^1$ = [-1,\\ -2, $-\frac{\pi}{8}$, 0.0, -0.2] and $x_{max}^1$ = [1, 2, $\frac{\pi}{8}$, 1.0, 0.2]. The follower agents' initial positions were placed in the wedge formation with some noise sampled from a uniform distribution with a bound of $\pm$[0.5, 0.5, $\frac{\pi}{16}$, 0.5, 0.2]. 

 We used an open environment to test the inter-agent collision avoidance further. Three agents were placed symmetrically around a 5-meter-radius circle, given a goal point antipodal to the start position. In these trials, the optimal paths of all three agents would meet in the center of the circle to encourage head-on collisions. 
 
 The parameters were $\omega_1 = 0.5$, $\omega_2 = 0.1$, $k$ = 3, $k_{obs,static} = 1$, $k_{att,static} = 8$, $k_{obs,agent} = k_{att, agent} = 0.5$, ${\bf Q}_f^1 = diag([1.0 \ 1.0])$ and ${\bf Q}_f^{2,3} = diag([1.0 \ 1.5])$


\subsection{Centralized Model}

We tested the centralized model on the obstacle field, running 200 optimization iterations per planning cycle and 4096 trajectory samples per optimization iteration. The leader agent followed a path generated using the rapidly-exploring random trees (RRT) algorithm \cite{lavalle1998rapidly} from its initial position to the goal. The leader's RRT defined the follower's formation points. While the leader successfully navigated the obstacle field without collisions, the followers often became trapped in the obstacle field. The centralized model never violated its bounds on cost or constraints.

In the symmetric antipodal goal experiment, the centralized approach deconflicted its trajectories without inter-agent gyroscopic avoidance, allowing each agent to reach the goal without a collision.

\subsection{Distributed Model}

\begin{figure}
    \vspace{5pt}
    \centering
    \includegraphics[width=\linewidth]{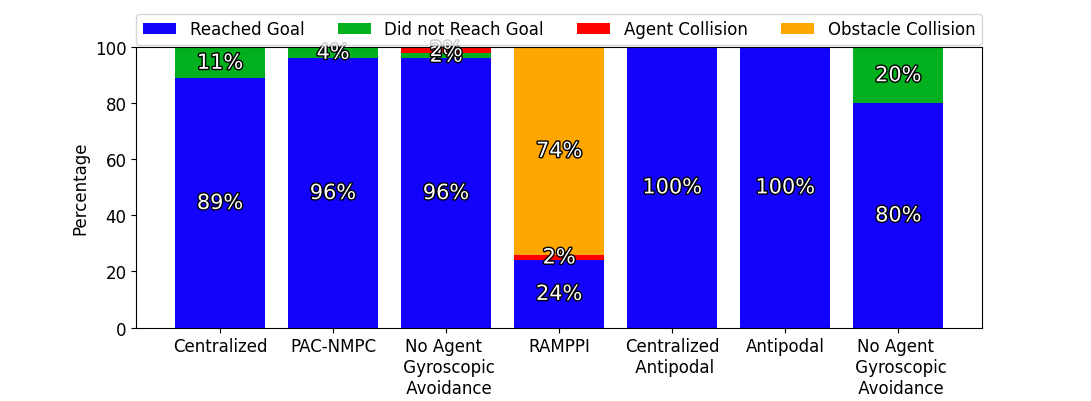}
    \caption{Trial Outcomes: With RA-MPPI, the followers struggled to keep up with the constantly changing cost function and collided with obstacles. The centralized model ran 19 Trials, and the antipodal test ran 10 trials; the rest ran 50 trials}
    \label{fig:CVDTO}
\end{figure}

The distributed model was tested at 200 iterations per planning cycle and 1024 trajectory samples per optimization iteration. The gyroscopic obstacle avoidance for inter-agent collisions was turned on and off in two separate trials. Figure \ref{fig:CVDTO} illustrates the outcomes, including whether a trial resulted in an obstacle crash, a collision with another agent, an agent getting trapped behind an obstacle, or all agents successfully reaching the goal.

The distributed approach can travel through the obstacle-filled environment in the same number of iterations but much quicker, with fewer instances of agents getting trapped in local minima compared to the centralized approach. This improvement is likely due to the increased complexity of the optimization problem as the number of control inputs increases in the centralized model.

\begin{figure}
    \vspace{5pt}
    \centering
    \includegraphics[width=\linewidth]{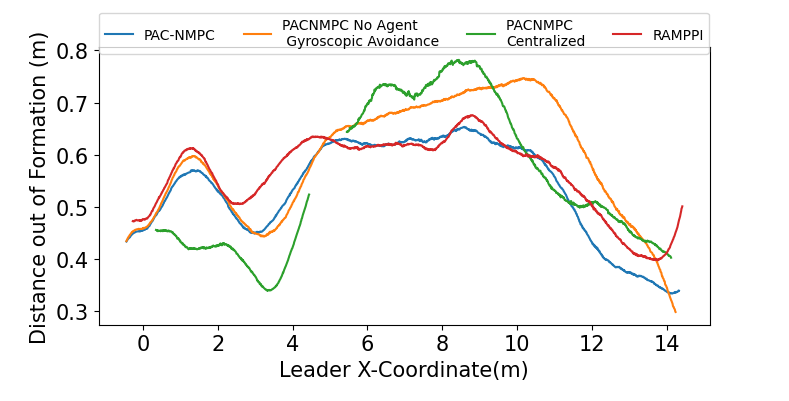}
    \caption{Centralized vs Distributed Formation Error: The centralized approach (green) held the best formation before the obstacles. As the agents encounter obstacles, the centralized approach exhibits a higher formation error due to frequent agent trapping.}
    \label{fig:CVDFE}
    \vspace{-10pt}
\end{figure}

The trials with and without the inter-agent gyroscopic avoidance led to a similar number of successful trials. However, the trials without led to poorer formation control within the obstacle field, as seen in Figure \ref{fig:CVDFE}. Not having gyroscopic avoidance led to agents avoiding collisions using a stop-and-go strategy, negatively impacting formations.

In the symmetric antipodal goal experiment, the distributed controller without inter-agent gyroscopic avoidance got stuck in deadlocks 20\% of the time, where the three agents avoided the collision but could not move towards the goal. When gyroscopic avoidance was enabled, these deadlocks were prevented, allowing a performance similar to that of the centralized controller.

\begin{figure}
    \centering
    \vspace{6pt}
    \includegraphics[width=\linewidth]{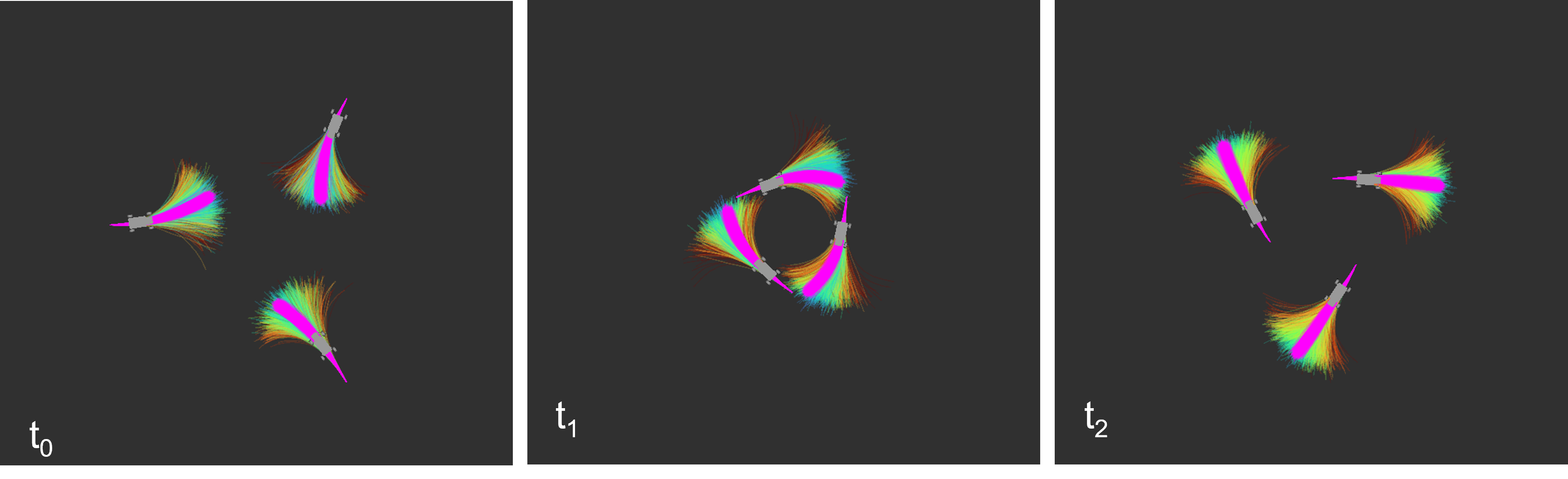}
    \caption{Symmetric Antipodal Experiment with Trajectory Deconflicting using Terminal Gyroscopic Obstacle Avoidance Cost}
    \label{fig:SYETD}
    \vspace{-10pt}
\end{figure}

\subsection{PAC-NMPC and RA-MPPI}
\begin{figure}
    \centering
    \vspace{-10pt}
    \includegraphics[width=\linewidth]{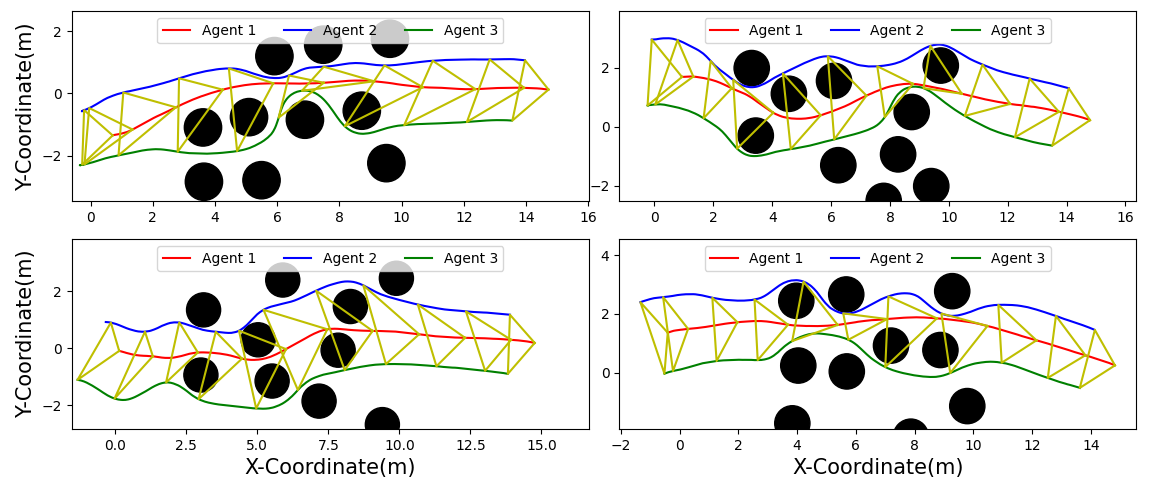}
    \caption{PAC-NMPC Formation}
    \label{fig:PAC12}
\end{figure}

\begin{figure}
    \vspace{5pt}
    \centering
    \includegraphics[width=\linewidth]{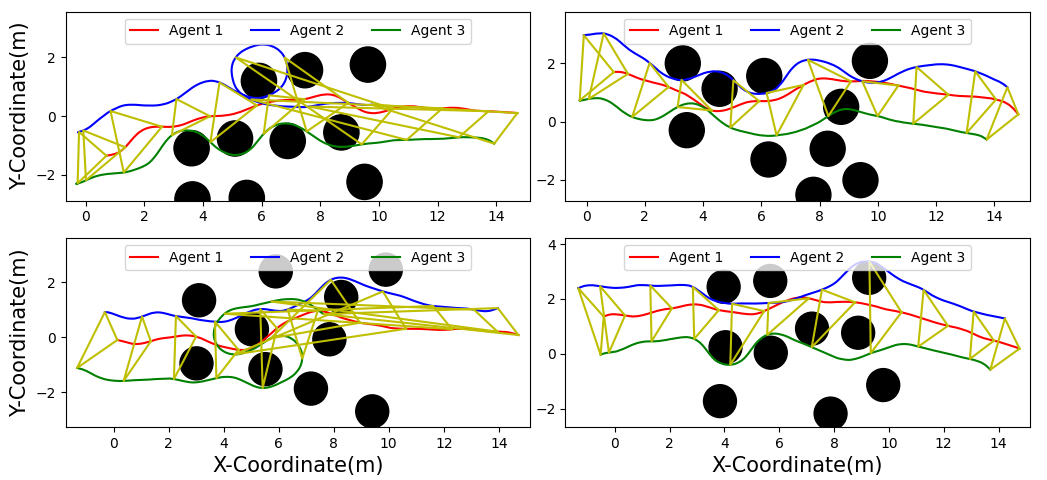}
    \caption{RA-MPPI Formation}
    \label{fig:RAMPPI12}
    \vspace{-10pt}
\end{figure}

We conducted additional trials with Risk Aware Model Predictive Path Integral (RA-MPPI) \cite{RA-MPPI}. Figure \ref{fig:CVDFE} reveals RA-MPPI resulted in a slightly higher average formation error than PAC-NMPC. The followers struggled to avoid crashes with obstacles, as seen in Figure \ref{fig:CVDTO}. Figures \ref{fig:PAC12} and \ref{fig:RAMPPI12} show the trajectories through the environment, where we can see RA-MPPI cutting through obstacles as formation points change or getting turned around to avoid collisions. PAC-NMPC was able to keep the wedge formation through the same obstacle fields. Furthermore, PAC-NMPC performed with a lower probability of constraint violations in the follower's planning than RA-MPPI. However, PAC-NMPC did run 23\% slower and the centralized approach ran 453\% slower per planning cycle, than RA-MPPI.

\subsection{Impact of State Measurement Uncertainty}
\begin{figure}
    \vspace{5pt}
    \centering
    \includegraphics[width=\linewidth]{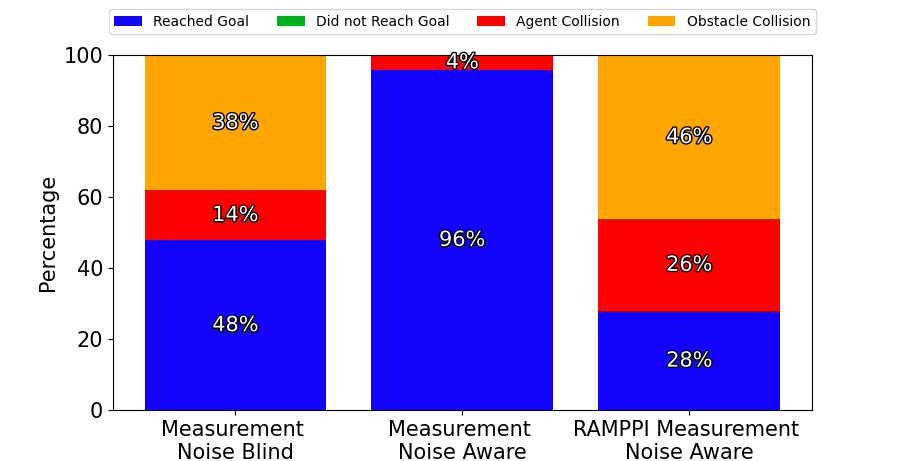}
    \caption{Measurement Noise Outcomes (50 Trials): The noise-aware approach significantly reduced agent collisions by 3.5 times at a covariance of noise of 0.9 meters}
    \label{fig:NTO}
    \vspace{-5pt}
\end{figure}

The trials were run with uncertain state positions, where the shared states were drawn from a normal distribution with a covariance of 0.9 meters in the x- and y-coordinates. As shown in Figure \ref{fig:NTO}, PAC-NMPC experienced a significant increase in collisions with agents and obstacles when state measurement noise was not accounted for. These static obstacle collisions often occurred as the noise-blind controller attempted to avoid an agent that was not there, leading to constraint violations in nearly all sampled trajectories. 

In contrast, the measurement noise-aware model successfully completed the same number of trials as the trial with ground truth measurements. While it did not eliminate collisions, it significantly reduced their frequency compared to the noise-blind case, preventing the agents from falling out of formation or colliding with each other as often. The noise blind controller also violated its cost bound in $\approx$10\% of plans, whereas the noise aware controller violated cost bounds $\approx$2\% of the time.

\begin{figure}
    \centering
    \vspace{-15pt}
    \includegraphics[width=\linewidth]{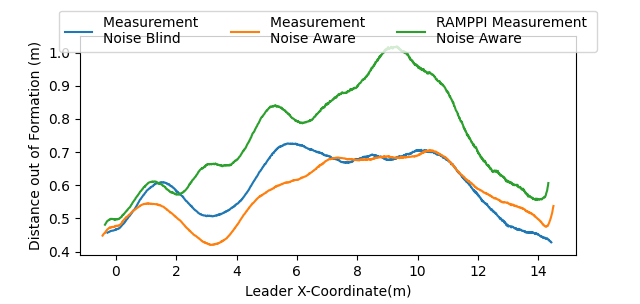}
    \caption{Noise vs Formation Error: RA-MPPI under the noise-aware controller (Green) held the worse formation control. PAC-NMPC, both noise-blind and -aware, maintained a more stable formation.}
    \label{fig:NFE}
\end{figure}

RA-MPPI struggled to compensate for the state uncertainties as effectively as PAC-NMPC, resulting in collision frequency on par with the noise-blind PAC-NMPC configuration. RA-MPPI also showed much worse formation performance over the trials, as seen in Figure \ref{fig:NFE}. Table \ref{tab:NT} records the average constraint probability, showing that RA-MPPI had a higher average constraint violation, and the followers also had a higher probability of constraint violation than the leaders.

\begin{table}
  
  \centering
  \vspace{8pt}
  \begin{tabular}{||c| c c c||}
    \hline
    \textbf{\thead{Trial Type}}  & \textbf{\thead{Number of\\ Collisions}} & \textbf{\thead{ Leader \\Constraint }} & \textbf{\thead{ Follower \\Constraint}}\\ [0.5ex]
    \hline\hline
    \thead{PAC-NMPC} & 0 & 0.00465 & 0.013 \\
    \hline
    \thead{RA-MPPI} & 38 & 0.0521 & 0.1788\\ [1ex]
    \hline
    \thead{PAC-NMPC State \\Noise Blind} & 26 & 0.1208 & 0.148\\
    \hline
    \thead{PAC-NMPC State \\Noise Aware} & 2 & 0.0987 & 0.1854\\
    \hline
    \thead{RA-MPPI State \\Noise Aware} & 33 & 0.208 & 0.289\\ [1ex]
    \hline
  \end{tabular}
  \caption{Collisions and Average Constraint Probability}
  \label{tab:NT}
  \vspace{-20pt}
\end{table}

\subsection{Formation Points}
The trials were repeated using three different methods for calculating the followers' formation points: they were defined directly from the leader's RRT, calculated based on the final state of the leader's mean trajectory, or calculated by the follower using its own RRT, with the goal point defined by the leader's RRT.

\begin{figure}
    \centering
    \includegraphics[width=\linewidth]{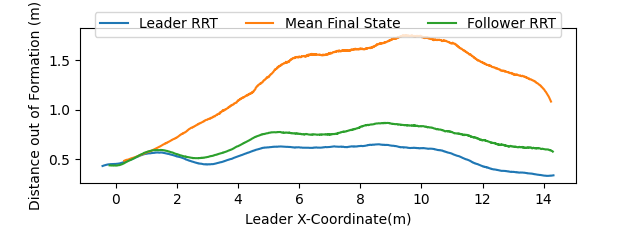}
    \caption{Formation Point Generation vs Formation Error: The RRT approaches (Blue, Green) lead to better formations as the mean final state (Orange) often had agents make U-Turns at the start to get into formation}
    \label{fig:FPFE}
    \vspace{-10pt}
\end{figure}

\begin{figure}
    \vspace{-5pt}
    \centering
    \includegraphics[width=\linewidth]{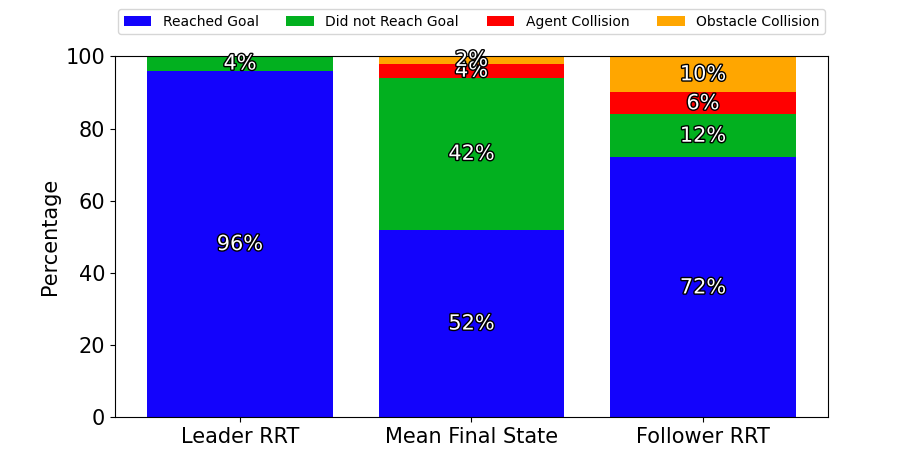}
    \caption{Formation Point Generation Method Outcomes (50 Trials): The follower RRT approach led to more collisions and local minima due to the RRT only accounting for agent position.}
    \label{fig:FETO}
    \vspace{-5pt}
\end{figure}

Formation point calculation methods based on the leader's RRT generally outperformed those using the mean final state in both successful trial percentage and formation error as seen in Figures \ref{fig:FPFE} and \ref{fig:FETO}. The follower RRT-based approach resulted in more agents trapped behind obstacles than when the formation point is only defined by the leader's RRT. 


\subsection{Symmetric Antipodal Under State Uncertainty}

\begin{figure}
    \vspace{5pt}
    \centering
    \includegraphics[width=\linewidth]{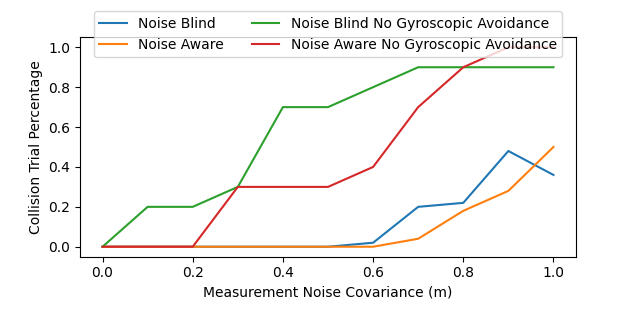}
    \caption{Collision Percentage vs Measurement Covariance. The noise-aware state sampling and gyroscopic avoidance increase the controller's ability to avoid collisions.}
    \label{fig:CoVN}
    \vspace{-5pt}
\end{figure}

We ran fifty trials in the symmetric antipodal setup with measurement noise variance from 0 to 1.0 meters. Figure \ref{fig:CoVN} shows that gyroscopic avoidance assists with avoiding collisions under measurement noise. The trials without the gyroscopic avoidance began to collide at 0.1 meters with the noise-blind controller but at 0.2 meters with the noise-aware controller. However, adding the gyroscopic avoidance increased the collision avoidance to 0.6 meters in the noise-blind case and 0.7 meters in the noise-aware case. The noise-aware controller reduced the number of collisions as the noise increased and achieved fewer constraint violations than the noise-blind controller. 

\subsection{Extension to Higher Dimensions}
\begin{figure}
    \centering
    \includegraphics[width=\linewidth]{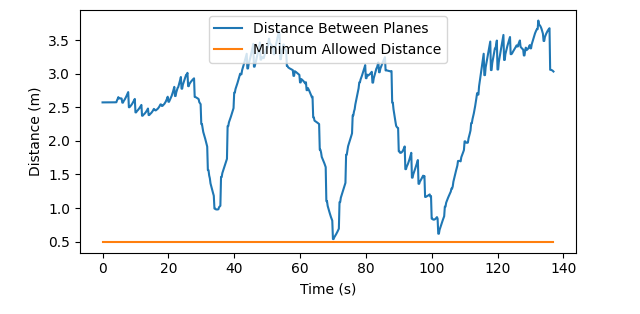}
    \caption{The Distance between Two Fixed-Wing Planes. During the two full loops, the planes never violated the safety radius}
    \label{fig:FWP}
    \vspace{-8pt}
\end{figure}

Two fixed-wing planes were simulated following the same path in opposite directions. The fixed wing is described by a complex nonlinear dynamics model with a 17-dimension state and four control inputs \cite{basescu2020directnmpcpoststallmotion}. PAC-NMPC avoided collisions at 100 iterations using 3D gyroscopic avoidance. Figure \ref{fig:FWP} shows the distance between planes as they follow a rectangular path in opposite directions. 


\section{Discussion}
In this paper, we presented a distributed receding-horizon SNMPC approach for navigating a multi-robot team through an obstacle field in formation. We demonstrated in simulation that using a gyroscopic avoidance-based cost enables the controller to keep formation by avoiding collisions and local minima. Further, we showed that control policy sharing could enable collision avoidance under state measurement uncertainty. Future work could explore heterogeneous teams and apply this model to more complex environments and dynamics.

\section*{ACKNOWLEDGMENT}
We gratefully acknowledge the support of the Army Research Laboratory under grant W911NF-22-2-0241. 



\bibliographystyle{IEEEtran}
\clearpage
\bibliography{references}

\begin{thebibliography}{10}
\providecommand{\url}[1]{#1}
\csname url@samestyle\endcsname
\providecommand{\newblock}{\relax}
\providecommand{\bibinfo}[2]{#2}
\providecommand{\BIBentrySTDinterwordspacing}{\spaceskip=0pt\relax}
\providecommand{\BIBentryALTinterwordstretchfactor}{4}
\providecommand{\BIBentryALTinterwordspacing}{\spaceskip=\fontdimen2\font plus
\BIBentryALTinterwordstretchfactor\fontdimen3\font minus \fontdimen4\font\relax}
\providecommand{\BIBforeignlanguage}[2]{{%
\expandafter\ifx\csname l@#1\endcsname\relax
\typeout{** WARNING: IEEEtran.bst: No hyphenation pattern has been}%
\typeout{** loaded for the language `#1'. Using the pattern for}%
\typeout{** the default language instead.}%
\else
\language=\csname l@#1\endcsname
\fi
#2}}
\providecommand{\BIBdecl}{\relax}
\BIBdecl

\bibitem{stern2019multi}
R.~Stern, ``Multi-agent path finding--an overview,'' \emph{Artificial Intelligence: 5th RAAI Summer School, Dolgoprudny, Russia, July 4--7, 2019, Tutorial Lectures}, pp. 96--115, 2019.

\bibitem{Sandip2012}
S.~Kumar and S.~Chakravorty, ``Multi-agent generalized probabilistic roadmaps: Magprm,'' in \emph{2012 IEEE/RSJ International Conference on Intelligent Robots and Systems}, 2012, pp. 3747--3753.

\bibitem{Wu2021}
Y.~Wu, B.~Jiang, and H.~Xu, ``Formation control strategy of multi-agent system with improved probabilistic roadmap method applied in restricted environment,'' in \emph{2021 6th International Conference on Computational Intelligence and Applications (ICCIA)}, 2021, pp. 233--237.

\bibitem{Beyoglu2022}
B.~R. Hikmet~Beyoglu, Stephan~Weiss, ``Multi-agent path planning and trajectory generation for confined environments,'' \emph{International Conference on Unmanned Aircraft Systems}, 2022.

\bibitem{Liu2017}
\BIBentryALTinterwordspacing
X.~Liu, S.~S. Ge, and C.-H. Goh, ``Formation potential field for trajectory tracking control of multi-agents in constrained space,'' \emph{International Journal of Control}, vol.~90, no.~10, pp. 2137--2151, 2017. [Online]. Available: \url{https://doi.org/10.1080/00207179.2016.1237044}
\BIBentrySTDinterwordspacing

\bibitem{Polevoy2023}
A.~Polevoy, M.~Kobilarov, and J.~Moore, ``Probably approximately correct nonlinear model predictive control (pac-nmpc),'' \emph{IEEE Robotics and Automation Letters}, pp. 1--8, 2023.

\bibitem{Hai2021}
H.~T., H.~Do, H.~T. Hua, H.~T. Hua, H.~Hua, M.~T. Nguyen, M.~T. Nguyen, M.~T.~H. Nguyen, M.~T. Nguyen, C.~V. Nguyen, C.~Nguyen, C.~V. Nguyen, C.~Nguyen, H.~Q. Nguyen, H.~Nguyen, H.~Nguyen, H.~T. Nguyen, N.~K. Nguyen, N.~Nguyen, N.~T.~T. Nguyen, N.~T. Nguyen, and N.~T.~T. Nguyen, ``Formation control algorithms for multiple-uavs: A comprehensive survey,'' \emph{EAI Endorsed Trans. Ind. Networks Intell. Syst.}, 2021.

\bibitem{Chen_2023}
Q.~jie Chen, Y.~Wang, Y.~Jin, T.~Wang, X.~Nie, and T.~Yan, ``A survey of an intelligent multi-agent formation control,'' \emph{Applied Sciences}, 2023.

\bibitem{Balch_1998}
T.~Balch, T.~Balch, R.~C. Arkin, and R.~C. Arkin, ``Behavior-based formation control for multirobot teams,'' \emph{null}, 1998.

\bibitem{Lee_2018}
G.~Lee, G.~Lee, D.~Chwa, D.~Chwa, and D.~Chwa, ``Decentralized behavior-based formation control of multiple robots considering obstacle avoidance,'' \emph{Intelligent Service Robotics}, 2018.

\bibitem{Emile_2020}
M.~B. Emile, M.~B. Emile, O.~M. Shehata, O.~M. Shehata, A.~El-Badawy, and A.~A. El-Badawy, ``A decentralized control of multiple unmanned aerial vehicles formation flight considering obstacle avoidance,'' \emph{International Conference on Control, Mechatronics and Automation}, 2020.

\bibitem{ren2006consensus}
W.~Ren, ``Consensus based formation control strategies for multi-vehicle systems,'' in \emph{2006 American Control Conference}.\hskip 1em plus 0.5em minus 0.4em\relax IEEE, 2006, pp. 6--pp.

\bibitem{Xiao_2009}
F.~Xiao, F.~Xiao, F.~Xiao, F.~Xiao, L.~Wang, L.~Wang, J.~Chen, J.~Chen, J.~Chen, J.~Chen, J.~Chen, Y.~Gao, and Y.~Gao, ``Brief paper: Finite-time formation control for multi-agent systems,'' \emph{Automatica}, 2009.

\bibitem{saber2003flocking}
R.~O. Saber and R.~M. Murray, ``Flocking with obstacle avoidance: Cooperation with limited communication in mobile networks,'' in \emph{42nd IEEE International Conference on Decision and Control (IEEE Cat. No. 03CH37475)}, vol.~2.\hskip 1em plus 0.5em minus 0.4em\relax IEEE, 2003, pp. 2022--2028.

\bibitem{Wang_2014}
W.~Wang, W.~Wang, W.~Wang, W.~Wang, J.~Huang, J.~Huang, W.~Chen, C.~Wen, H.~Fan, and H.~Fan, ``Distributed adaptive control for consensus tracking with application to formation control of nonholonomic mobile robots,'' \emph{Automatica}, 2014.

\bibitem{Romero_2023}
J.~G. Romero, E.~Nuño, E.~Restrepo, and I.~Sarras, ``Global consensus-based formation control of nonholonomic mobile robots with time-varying delays and without velocity measurements,'' \emph{IEEE Transactions on Automatic Control}, 2023.

\bibitem{Khatib1990}
\BIBentryALTinterwordspacing
O.~Khatib, \emph{Real-Time Obstacle Avoidance for Manipulators and Mobile Robots}.\hskip 1em plus 0.5em minus 0.4em\relax New York, NY: Springer New York, 1990, pp. 396--404. [Online]. Available: \url{https://doi.org/10.1007/978-1-4613-8997-2_29}
\BIBentrySTDinterwordspacing

\bibitem{Colledanchise_2013}
M.~Colledanchise, M.~Colledanchise, D.~V. Dimarogonas, D.~V. Dimarogonas, P.~Ögren, and P.~Ögren, ``Obstacle avoidance in formation using navigation-like functions and constraint based programming,'' \emph{2013 IEEE/RSJ International Conference on Intelligent Robots and Systems}, 2013.

\bibitem{Liu_2017}
X.~Liu, X.~Liu, S.~S. Ge, S.~S. Ge, C.~H. Goh, and C.-H. Goh, ``Formation potential field for trajectory tracking control of multi-agents in constrained space,'' \emph{International Journal of Control}, 2017.

\bibitem{Sun_2020}
Y.~Sun, Y.~Sun, X.~Hu, X.~Hu, J.~Xiao, J.~Xiao, J.~Xiao, G.~Zhang, G.~Zhang, G.~Zhang, S.~Wang, S.~Wang, L.~Liu, and L.~Liu, ``Multi-agent cluster systems formation control with obstacle avoidance,'' \emph{2020 15th IEEE Conference on Industrial Electronics and Applications (ICIEA)}, 2020.

\bibitem{Tan_1996}
K.-H. Tan, K.-H. Tan, M.~A. Lewis, and M.~Lewis, ``Virtual structures for high-precision cooperative mobile robotic control,'' \emph{Proceedings of IEEE/RSJ International Conference on Intelligent Robots and Systems. IROS '96}, 1996.

\bibitem{Lewis_1997}
M.~A. Lewis, M.~A. Lewis, K.-H. Tan, and K.-H. Tan, ``High precision formation control of mobile robots using virtual structures,'' \emph{Autonomous Robots}, 1997.

\bibitem{Wu_2021}
Y.~Wu, Y.~Wu, B.~Jiang, B.~Jiang, H.~Xu, and H.~Xu, ``Formation control strategy of multi-agent system with improved probabilistic roadmap method applied in restricted environment,'' \emph{International Conference on Computer and Information Application}, 2021.

\bibitem{Beckers_2021}
T.~Beckers, T.~Beckers, S.~Hirche, S.~Hirche, L.~Colombo, and L.~Colombo, ``Online learning-based formation control of multi-agent systems with gaussian processes,'' \emph{IEEE Conference on Decision and Control}, 2021.

\bibitem{Lima_2021}
J.~V. C.~F. de~Lima, J.~F. de~Lima, E.~M. Belo, E.~M. Belo, V.~A. da~Silva~Marques, V.~A. da~Silva~Marques, and V.~A. da~Silva~Marques, ``Multi-agent path planning with nonlinear restrictions,'' \emph{Evolutionary Intelligence}, 2021.

\bibitem{Ryou_2022}
G.~Ryou, E.~Tal, and S.~Karaman, ``Cooperative multi-agent trajectory generation with modular bayesian optimization,'' \emph{Robotics: Science and Systems}, 2022.

\bibitem{Obradović_2023}
J.~Obradović, M.~Križmančić, and S.~Bogdan, ``Decentralized multi-robot formation control using reinforcement learning,'' \emph{International Symposium on Information, Communication and Automation Technologies}, 2023.

\bibitem{Xue_2021}
J.~Xue, X.~Kong, B.~Dong, and M.~Xu, ``Multi-agent path planning based on mpc and ddpg,'' \emph{arXiv.org}, 2021.

\bibitem{Roldao_2014}
V.~Roldao, V.~Roldão, R.~Cunha, R.~Cunha, D.~Cabecinhas, D.~Cabecinhas, C.~Silvestre, C.~Silvestre, P.~Oliveira, and P.~Oliveira, ``A leader-following trajectory generator with application to quadrotor formation flight,'' \emph{Robotics and Autonomous Systems}, 2014.

\bibitem{Xiao_2016}
H.~Xiao, H.~Xiao, Z.~Li, Z.~Li, C.~L.~P. Chen, and C.~L.~P. Chen, ``Formation control of leader–follower mobile robots’ systems using model predictive control based on neural-dynamic optimization,'' \emph{IEEE Transactions on Industrial Electronics}, 2016.

\bibitem{Xiao_2017}
H.~Xiao, H.~Xiao, C.~L.~P. Chen, and C.~L.~P. Chen, ``Leader-follower multi-robot formation system using model predictive control method based on particle swarm optimization,'' \emph{Youth Academic Annual Conference of Chinese Association of Automation}, 2017.

\bibitem{Xiao_2019}
H.~Xiao, H.~Xiao, H.~Xiao, C.~L.~P. Chen, C.~L.~P. Chen, and C.~Chen, ``Leader-follower consensus multi-robot formation control using neurodynamic-optimization-based nonlinear model predictive control,'' \emph{IEEE Access}, 2019.

\bibitem{Xu_2023}
T.~Xu, J.~K. Liu, Z.~Zhang, G.~Chen, D.~Cui, and H.~Li, ``Distributed mpc for trajectory tracking and formation control of multi-uavs with leader-follower structure,'' \emph{IEEE Access}, 2023.

\bibitem{Luis_2020}
C.~E. Luis, M.~Vukosavljev, A.~P. Schoellig, and A.~P. Schoellig, ``Online trajectory generation with distributed model predictive control for multi-robot motion planning,'' \emph{IEEE Robotics and Automation Letters}, 2020.

\bibitem{Trevisan2024}
E.~Trevisan and J.~Alonso-Mora, ``Biased-mppi: Informing sampling-based model predictive control by fusing ancillary controllers,'' \emph{IEEE Robotics and Automation Letters}, vol.~9, no.~6, pp. 5871--5878, 2024.

\bibitem{tajbakhsh2024conflict}
A.~Tajbakhsh, L.~T. Biegler, and A.~M. Johnson, ``Conflict-based model predictive control for scalable multi-robot motion planning,'' in \emph{2024 IEEE International Conference on Robotics and Automation (ICRA)}.\hskip 1em plus 0.5em minus 0.4em\relax IEEE, 2024, pp. 14\,562--14\,568.

\bibitem{Nfaileh_2022}
N.~Nfaileh, N.~Nfaileh, K.~Alipour, K.~Alipour, B.~Tarvirdizadeh, B.~Tarvirdizadeh, A.~Hadi, and A.~Hadi, ``Formation control of multiple wheeled mobile robots based on model predictive control,'' \emph{Robotica}, 2022.

\bibitem{Satir_2023}
S.~Satir, Y.~F. Aktaş, S.~Atasoy, M.~M. Ankaralı, and E.~Sahin, ``Distributed model predictive formation control of robots with sampled trajectory sharing in cluttered environments,'' \emph{IEEE/RJS International Conference on Intelligent RObots and Systems}, 2023.

\bibitem{kobilarov2015sample}
M.~Kobilarov, ``Sample complexity bounds for iterative stochastic policy optimization,'' \emph{Advances in Neural Information Processing Systems}, vol.~28, 2015.

\bibitem{underactuated}
\BIBentryALTinterwordspacing
R.~Tedrake, \emph{Underactuated Robotics}, 2023. [Online]. Available: \url{https://underactuated.csail.mit.edu}
\BIBentrySTDinterwordspacing

\bibitem{Garimella2016}
G.~Garimella, M.~Sheckells, and M.~Kobilarov, ``A stabilizing gyroscopic obstacle avoidance controller for underactuated systems,'' in \emph{2016 IEEE 55th Conference on Decision and Control (CDC)}, 2016, pp. 5010--5016.

\bibitem{chang2003}
D.~E. Chang and J.~E. Marsden, ``Gyroscopic forces and collision avoidance with convex obstacles,'' in \emph{New Trends in Nonlinear Dynamics and Control and their Applications}, W.~Kang, C.~Borges, and M.~Xiao, Eds.\hskip 1em plus 0.5em minus 0.4em\relax Berlin, Heidelberg: Springer Berlin Heidelberg, 2003, pp. 145--159.

\bibitem{lavalle1998rapidly}
S.~M. LaValle, ``Rapidly-exploring random trees: A new tool for path planning,'' 1998.

\bibitem{RA-MPPI}
J.~Yin, Z.~Zhang, and P.~Tsiotras, ``Risk-aware model predictive path integral control using conditional value-at-risk,'' 09 2022.

\bibitem{basescu2020directnmpcpoststallmotion}
\BIBentryALTinterwordspacing
M.~Basescu and J.~Moore, ``Direct nmpc for post-stall motion planning with fixed-wing uavs,'' 2020. [Online]. Available: \url{https://arxiv.org/abs/2001.11478}
\BIBentrySTDinterwordspacing

\end{thebibliography}


\end{document}